\documentclass[a4paper,fleqn]{cas-dc}

\usepackage[numbers]{natbib}
\usepackage{lineno,hyperref}
\usepackage{amsthm}
\usepackage{amsmath}
\usepackage{subfigure}
\usepackage{algorithm}
\usepackage[noend]{algpseudocode}
\usepackage{multirow}

\newtheorem{mht_def}{Definition} 
\modulolinenumbers[5]

\DeclareMathOperator{\msd}{msd}
\DeclareMathOperator{\mse}{mse}
\DeclareMathOperator{\Kruskal}{Kruskal}

\def\tsc#1{\csdef{#1}{\textsc{\lowercase{#1}}\xspace}}
\tsc{WGM}
\tsc{QE}
\tsc{EP}
\tsc{PMS}
\tsc{BEC}
\tsc{DE}
\begin{document}
\let\WriteBookmarks\relax
\def\floatpagepagefraction{1}
\def\textpagefraction{.001}
\shorttitle{Graph-Based Parallel Large Scale Structure from Motion}
\shortauthors{Yu Chen et~al.}

\title [mode = title]{Graph-Based Parallel Large Scale Structure from Motion}

\author[pku]{Yu Chen}%[type=editor,
                     %   auid=000,bioid=1,
                      %  prefix=Sir,
                     %   role=Researcher,
                     %   orcid=0000-0001-7511-2910]
%\cormark[1]
%\fnmark[1]
\ead{rainyChen@pku.edu.cn}
%\ead[url]{www.cvr.cc, cvr@sayahna.org}
%\credit{Conceptualization of this study, Methodology, Software}

\author[nlpr, ucas]{Shuhan Shen}[style=chinese]
\ead{shshen@nlpr.ia.ac.cn}

\author[pku]{Yisong Chen}
\ead{chenyisong@pku.edu.cn}

\author[pku]{Guoping Wang}
\ead{wgp@pku.edu.cn}

%\credit{Data curation, Writing - Original draft preparation}

\address[pku]{Graphics and Interactive Lab, Department of Computer Science and Technology, School of Electronic Engineering and Computer Science, Peking University}
\address[nlpr]{National Laboratory of Pattern Recognition, Institute of Automation, Chinese Academy of Sciences, Beijing 100190, China}
\address[ucas]{University of Chinese Academy of Sciences, Beijing 100049, China}
%\cortext[cor1]{Corresponding author}
%\cortext[cor2]{Principal corresponding author}
%\fntext[fn1]{This is the first author footnote. but is common to third
%  author as well.}
%\fntext[fn2]{Another author footnote, this is a very long footnote and
%  it should be a really long footnote. But this footnote is not yet
%  sufficiently long enough to make two lines of footnote text.}

%\nonumnote{This note has no numbers. In this work we demonstrate $a_b$
%  the formation Y\_1 of a new type of polariton on the interface
%%  between a cuprous oxide slab and a polystyrene micro-sphere placed
%  on the slab.
%  }

\begin{abstract}
While Structure from Motion achieves great success in 3D reconstruction, it still meets challenges on large scale scenes. Incremental SfM approaches are robust to outliers, but are limited by low efficiency and easy suffer from drift problem. Though Global SfM methods are more efficient than incremental approaches, they are sensitive to outliers, and would also meet memory limitation and time bottleneck. In this work, large scale SfM is deemed as a graph problem, where graph are respectively constructed in image clustering step and local reconstructions merging step. By leveraging the graph structure, we are able to handle large scale dataset in divide-and-conquer manner. Firstly, images are modelled as graph nodes, with edges are retrieved from geometric information after feature matching. Then images are divided into into independent clusters by a image clustering algorithm, and followed by a subgraph expansion step, the connection and completeness of scenes are enhanced by walking along a maximum spanning tree, which is utilized to construct overlapping images between clusters. Secondly, Image clusters are distributed into servers to execute SfM in parallel mode. Thirdly, after local reconstructions complete, we construct a minimum spanning tree to find accurate similarity transformations. Then the minimum spanning tree is transformed into a Minimum Height Tree to find a proper anchor node, and is further utilized to prevent error accumulation. We evaluate our approach on various kinds of datasets and our approach shows superiority over the state-of-the-art in accuracy and efficiency. Our algorithm is open-sourced in \href{GraphSfM}{https://github.com/AIBluefisher/GraphSfM}.
\end{abstract}

%\begin{graphicalabstract}
%\includegraphics{figs/grabs.pdf}
%\end{graphicalabstract}

%\begin{highlights}
%\item Research highlights item 1
%\item Research highlights item 2
%\item Research highlights item 3
%\end{highlights}


\maketitle

\section{Introduction}

\label{sec:intro}
The study of Structure-from-Motion (SfM) has made rapid progress in recent years. It has achieved great success in small to medium scale scenes. However, reconstructing large scale datasets remains a big challenge in terms of both efficiency and robustness.

Since \cite{DBLP:conf/iccv/AgarwalSSSS09} has achieved great success and has become a milestone, incremental approaches have been widely used in modern SfM applications\cite{DBLP:conf/cvpr/SnavelySS08, DBLP:conf/3dim/Wu13, DBLP:conf/accv/MoulonMM12, DBLP:conf/mm/SweeneyHT15, DBLP:conf/cvpr/SchonbergerF16, DBLP:journals/pr/GaoHCSH18, DBLP:journals/pr/LiuTS20}. The geometric filtering combined with RANSAC \cite{DBLP:journals/cacm/FischlerB81} process can remove outliers effectively. Starting with a robust initial seed reconstruction, incremental SfM then adds camera one by one by PnP \cite{DBLP:conf/cvpr/KneipSS11, DBLP:journals/ijcv/LepetitMF09}. After cameras are registered successfully, an additional bundle adjustment step is used to optimize both poses and 3D points \cite{DBLP:conf/iccvw/TriggsMHF99}, which makes incremental SfM robust and accurate. However, incremental SfM also becomes inefficient and would meet memory bottleneck on large scale datasets due to the repetitive optimization step. Besides, the manner of adding new views incrementally makes these kinds of approaches suffer from drift easily, though an additional re-triangulation step is used \cite{DBLP:conf/3dim/Wu13}.

Global SfM approaches \cite{DBLP:conf/cvpr/Govindu01, DBLP:conf/eccv/WilsonS14, DBLP:journals/pami/CrandallOSH13, DBLP:conf/iccv/CuiT15, DBLP:conf/iccv/ChatterjeeG13, DBLP:conf/eccv/HavlenaTP10, DBLP:conf/eccv/WilsonBS16, Govindu2006Robustness, Govindu2004Lie, DBLP:conf/cvpr/OzyesilS15, Moulon2013Global} have advantages over incremental ones in efficiency. When all available relative motions are obtained, global approaches first obtain global rotations by solving the rotation averaging problem efficiently and robustly \cite{DBLP:conf/cvpr/Govindu01, DBLP:conf/cvpr/Govindu04, DBLP:journals/ijcv/HartleyTDL13, DBLP:conf/cvpr/HartleyAT11, DBLP:conf/iccv/ChatterjeeG13, DBLP:journals/pami/ChatterjeeG18, DBLP:conf/cvpr/ErikssonOKC18, DBLP:conf/eccv/WilsonBS16}. Then, global orientations and relative translations are used to estimate camera translations(or camera centers) by translation averaging \cite{DBLP:conf/eccv/WilsonS14, DBLP:conf/cvpr/OzyesilS15, DBLP:conf/eccv/GoldsteinHLVS16, DBLP:journals/corr/abs-1901-00643}. With known camera poses, triangulation(re-triangulation might be required) can be performed to obtain 3D points and then only once bundle adjustment step is required. Though global approaches are efficiency, the shortcomings are obviously: translation averaging is hard to solve, as relative translations only decode the direction of translation and the scale is unknown; outliers are still a head-scratching problem for translation averaging, which is the main reason that prohibit the practical use of global SfM approaches. %What is more, global SfM is still impractical when optimizing all variables at the final bundle adjustment \cite{DBLP:conf/iccvw/TriggsMHF99} step. 

To overcome the inefficiency problem in incremental SfM while to remain the robustness of reconstruction at the same time, a natural idea
is to do reconstruction in a divide-and-conquer manner. A pioneer work that proposed this idea is \cite{DBLP:conf/accv/BhowmickPCGB14} where images are first partitioned by graph cut and each sub-reconstruction is stitched by similarity transformation. Then followed by \cite{Zhu2017Parallel, DBLP:conf/cvpr/ZhuZZSFTQ18} where both the advantages of incremental and global approaches are utilized in each sub-reconstruction. However, both these divide-and-conquer approaches are more focused on the local reconstructions and their pipelines are lack of global consideration, which designed the clustering step and merging step independently, thus may lead to the failure of SfM.

Inspired by these previous outstanding divide-and-conquer work \cite{DBLP:conf/accv/BhowmickPCGB14, DBLP:conf/3dim/SweeneyFHT16, Zhu2017Parallel, DBLP:conf/cvpr/ZhuZZSFTQ18}, we solve large scale SfM in a parallel mode while the whole pipeline is designed with a unified framework based on graph theory. We claim the novelties of the proposed framework are: (1) The image clustering algorithm, where we first cluster images in different groups, then further expand image clusters by walking along a MaxST. The image clustering allows the distribution of local SfM tasks. (2) The final  local reconstructions merging algorithm, local reconstructions are accurately registered into a anchor node, by leveraging a MinST and a MHT, where the most accurate similarity transformations can be selected by the former, and the proper anchor node can be found by the latter. %The proposed novelty framework starts from a global perspective where the image clustering step is designed for both robust local reconstruction and final sub-reconstructions merging step, in which each cluster is deemed as a node inside a graph. And the sub-reconstructions merging step can further utilize the cluster graph structure to obtain robust fusing results.  
%More 
Specifically, first, images are divided into clusters with no overlap and each cluster is a graph node. Second, lost edges are collected and used to construct a maximum spanning tree (MaxST). Then, these lost edges are added along the MaxST to construct overlapped images and enhance the connections between clusters. Third, local SfM solvers are executed in parallel or distributed mode. At last, after all local SfM jobs finish, a novel sub-reconstructions merging algorithm is proposed for clusters registering. The most accurate $N - 1$ similarity transformations are selected within a minimum spanning tree (MinST) and a minimum height tree (MHT) is constructed to find a suitable reference frame and suppress the accumulated error.

Our contributions are mainly three folds: 
\begin{itemize}
    \item We proposed an robust image clustering algorithm, where images are clustered into groups of suitable size with overlap, the connectivity is enhanced with the help of an MaxST.
    \item We proposed a novel graph-based sub-model merging algorithm, where MinST is constructed to find accurate similarity transformations, and MHT is constructed to avoid error accumulation during the merging process.
    \item The time complexity is linearly related to the number of images, while most state-of-the-art algorithms are quadratic.
\end{itemize}

%-------------------------------------- Related Work -----------------------------------
\section{Related Work}

Large scale SfM becomes popular since \cite{DBLP:conf/iccv/AgarwalSSSS09, DBLP:conf/eccv/WilsonS14}, where they used unordered internet images as input, and utilized skeletal graph~\cite{DBLP:conf/cvpr/SnavelySS08, DBLP:conf/eccv/ShenZFZQ16} to avoid exhaustive feature matching~\cite{DBLP:conf/iccv/WangYY19, DBLP:journals/pami/YanCZYC16, DBLP:journals/tip/YanWZYC15}.

Some exciting work in large scale reconstruction is hierarchical SfM approaches \cite{Farenzena2009Structure, Gherardi2010Improving, Toldo2015Hierarchical, DBLP:conf/3dim/NiD12, DBLP:journals/cviu/ChenCLSW17}. These approaches take each image as a leaf node. Point clouds and camera poses are merged from bottom to top. The principle of "smallest first" is adopted to produce a balanced dendrogram, which makes hierarchical approaches insensitive to initialization and drift error. However, due to insufficient feature matching~\cite{DBLP:journals/ijcv/Lowe04, DBLP:journals/pr/MaJJG19}, the reconstructed scenes tend to lose scene details and become incomplete. Besides, the quality of reconstructions might be ruined by the selection of similar image pairs.

Some earlier work tries to solve large scale SfM via multi-cores \cite{DBLP:conf/iccv/AgarwalSSSS09}, or to reduce the burden of exhaustive pairwise matching by building skeletal graphs \cite{DBLP:conf/cvpr/SnavelySS08}. Bhowmick \cite{DBLP:conf/accv/BhowmickPCGB14} tried to solve large scale SfM in divide-and-conquer manner, and graph cut \cite{DBLP:journals/pami/DhillonGK07, DBLP:journals/pami/ShiM00} was adopted to do data partition. After all sub-reconstructions complete, additional cameras are registered in each sub-reconstruction to construct overlapping areas and then to fuse them. It was then improved in \cite{DBLP:conf/3dim/SweeneyFHT16} to cluster data set and merges each cluster by a distributed camera model \cite{DBLP:conf/eccv/SweeneyFHT14, DBLP:conf/3dim/SweeneyFHT16}. However, both \cite{DBLP:conf/3dim/SweeneyFHT16, DBLP:conf/accv/BhowmickPCGB14} either took no good consideration of the graph clustering strategy or neglected a careful design of clustering and merging algorithm, which makes reconstruction fragile and suffers from the drifting problem. Moreover, the loss of connections between different components makes the reconstruction fragile. Besides, the similarity score that is used as the weight in graph partition reduces the reliability of the result. One drawback that should be noticed is that the incremental merging process suffers from drifted errors, as well as traditional incremental approaches.

Follow the work of Bhowmick \cite{DBLP:conf/accv/BhowmickPCGB14}, \cite{Zhu2017Parallel, DBLP:conf/cvpr/ZhuZZSFTQ18} augment the graph cut process in \cite{DBLP:conf/accv/BhowmickPCGB14, DBLP:conf/3dim/SweeneyFHT16} by a two steps process: binary graph cut and graph expansion. In their work, the graph cut step and graph expansion step alternated and then converged when both size constraint and completeness constraint are satisfied. Then, components are registered by global motion averaging \cite{Zhu2017Parallel}. However, the translation averaging at the cluster level still suffers from outliers and may lead to disconnected models. This work was further improved in \cite{DBLP:conf/cvpr/ZhuZZSFTQ18}, where it adopted the clusters registering approach in \cite{DBLP:conf/accv/BhowmickPCGB14}, and then camera poses were divided into intra-cameras and inter-cameras for motion averaging, which can improve the convergence rate of final bundle adjustment \cite{DBLP:conf/iccvw/TriggsMHF99, DBLP:conf/cvpr/ErikssonBCI16, DBLP:conf/iccvw/RamamurthyLAPV17, DBLP:conf/iccv/ZhangZFQ17}.

%However, the two-steps image clustering algorithm is time-consuming, as it needs to traverse all discarded edges. Besides, the coarse alignment step of sub-reconstructions might make bundle adjustment hard to converge.

\section{Graph-Based Structure from Motion}

\begin{figure*}
\centering
\includegraphics[width=1.0\linewidth]{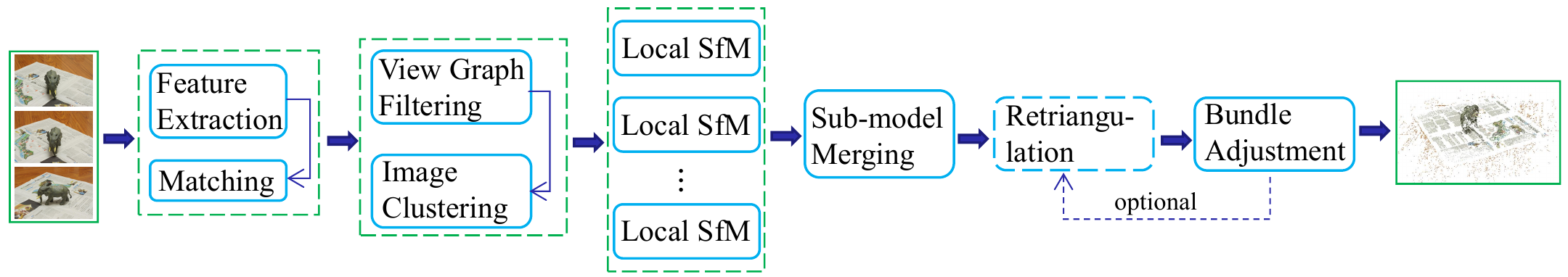}

\caption{Our proposed SfM pipeline. We first extract and match features. Epipolar geometries are estimated to filter matching outliers. After feature matching, we use our proposed image clustering algorithm to divide images into different groups. Then, clusters can be reconstructed by local SfM in parallel. After all local reconstructions are completed, we need to merge them together-as each local map has their own coordinate frame. And to avoid producing disconnected scenes after merging, all local reconstructions are merged with our graph-based merging algorithm. When we obtained a global map and camera poses, we can further re-triangulation more landmarks to recover more scene details. To minimize the reprojection error of the global map, a final bundle adjustment should be performed. The re-triangulation step and bundle adjustment step can be executed alternatively until convergence.}
\label{fig:pipeline_eval}
\end{figure*}

To deal with large scale datasets, we adopt the divide-and-conquer strategy that is similar to~\cite{DBLP:conf/accv/BhowmickPCGB14, Zhu2017Parallel}. For the sake of completeness and efficiency of reconstruction, we propose to use a unified graph framework to solve the image clustering and sub-reconstructions merging problem. The pipeline of our SfM algorithm is shown in Fig.\ref{fig:pipeline_eval}. Firstly, we extract features and use them for matching. Epipolar geometries are estimated to filter matching outliers. After feature matching, we use our proposed image clustering algorithm to divide images into different groups. Then, clusters can be reconstructed by local SfM in parallel. After all local reconstructions are completed, we need to merge them together-as each local map has their own coordinate frame. To avoid producing disconnected scenes after merging, all local reconstructions are merged with our graph-based merging algorithm. When we obtained a global map and camera poses, we can further re-triangulation more landmarks to recover more scene details. To minimize the reprojection error of the global map, a final bundle adjustment should be performed. The re-triangulation step and bundle adjustment step can be executed alternatively until convergence. In practice, it is enough to perform them once. We describe more details of our algorithm in the following subsections.

%%%%%%%%%%%%%%%%%%%% Figure %%%%%%%%%%%%%%%%%%%%%%%%%%%%%%%%%%%%%%%%%%%

\subsection{Image Clustering}
\label{sec:image_clustering}

We aim to group $m$ images into $k$ clusters, each cluster is under the memory limitation of the computer. Besides, each cluster should be reconstructed as accurately as possible, and not be influenced largely by the loss of geometric constraints. In this section, we share a simple but quite effective approach for a two steps image clustering: (1) Graph cutting. (2) Images expansion based on the maximum spanning tree. Both the two steps are based on the intuitive graph theory. Besides, we utilize two conditions proposed in \cite{Zhu2017Parallel} to constraint the clustering steps: size constraint and completeness constraint. Size constraint gives the upper bound of images in each cluster. Completeness constraint is defined as $\eta(i)=\frac{\sum_{j \neq i} \left |C_i \cap C_j \right |}{C_i}$, where $C_i, C_j$ respectively represents cluster $i, j$. Unlike the image clustering algorithm proposed in \cite{Zhu2017Parallel}, which alternates between graph cut and graph expansion, we just perform once graph cut and once subgraph expansion. And we claim the novelty of our expansion step is using a MaxST to assist the final fusing step.

\subsubsection{Image Clustering}
In graph cutting step, each image is deemed as a graph node, and each edge represents the connection between images. In the case of SfM, it can be represented by the results of two view geometries. The weight of edges is the number of matches after geometric filtering.
To solve this problem, the connection between images can be deemed as edges inside a graph. Suppose each camera is a graph node, the connection between two cameras can be deemed as a weighted edge(Here we referred to as \textbf{\textit{image edges}}). Consider the size constraint, each cluster should have a similar size. The \textit{image clustering} problem can be solved by graph cut \cite{DBLP:journals/pami/DhillonGK07, DBLP:journals/pami/ShiM00}. To enhance the connection of clusters and to align them together, an additional expansion step should be executed. In our case, we need to expand these independent clusters with some common images(we referred to as overlapping area), and then to compute similarity transformations to fuse them together. As the iterative approach of Zhu~\cite{Zhu2017Parallel} is time-consuming, we proposed our one step approach in expansion procedure in the next subsection.

\subsubsection{Subgraph Expansion}
In the subgraph expansion step, we generalize the graph into clusters level. Each cluster represents a node, the edges between clusters are the lost edges after graph cut (Here we refer as \textbf{\textit{cluster edges}}). First, we collect all the lost edges $E_{\text{lost}} = \{\mathcal{E}_{k_1 k_2, i}  | k_1, k_2 \in [K], i \in |\mathcal{E}_{k_1 k_2}|\}$ for cluster pairs, where $K$ is the number of clusters. Then, we construct a graph, the weight of the cluster edge is the number of lost image edges between two clusters. Intuitively, if there are more image edges are lost inside pairwise clusters, we prefer to construct connections for them to avoid the loss of information. With that in mind, once we obtain the cluster graph, a MaxST is constructed to induce the expansion step. We gather the image edges from the MaxST and sort them by descending order. We then add the lost image edges into clusters where completeness constraint is not satisfied, and only the top-$k$ edges are added. At last, we check all clusters and collect them together if the completeness constraint of any of them is not satisfied. For cluster edges that are not contained in the MaxST, we select them randomly and add the image edges into these clusters in a similar way.

\begin{figure}
\centering
\subfigure[]{
\begin{minipage}{0.45\linewidth}
\includegraphics[width=1.0\linewidth]{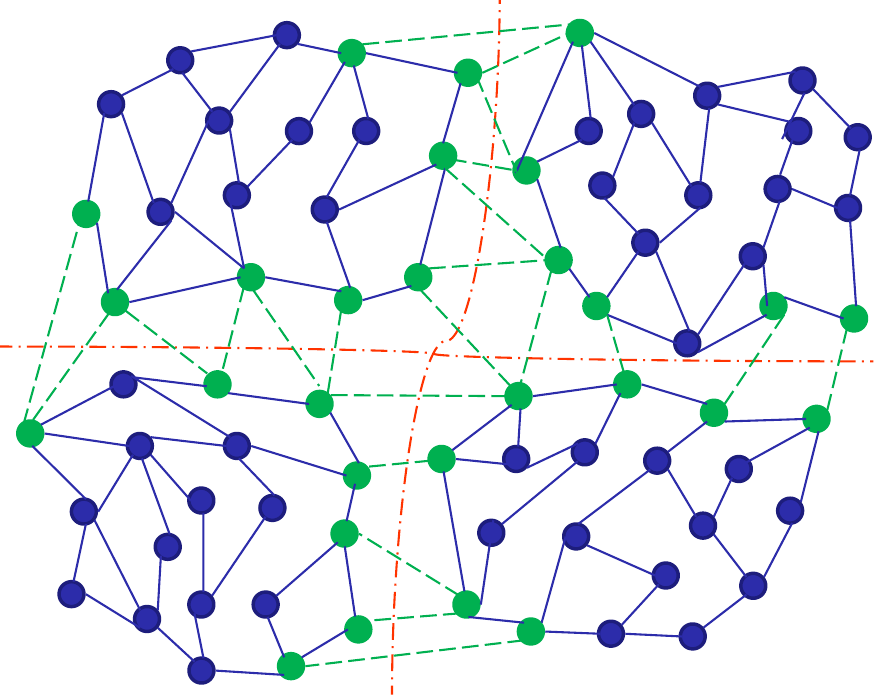}
\end{minipage}}
\subfigure[]{
\begin{minipage}{0.45\linewidth}
\includegraphics[width=1.0\linewidth]{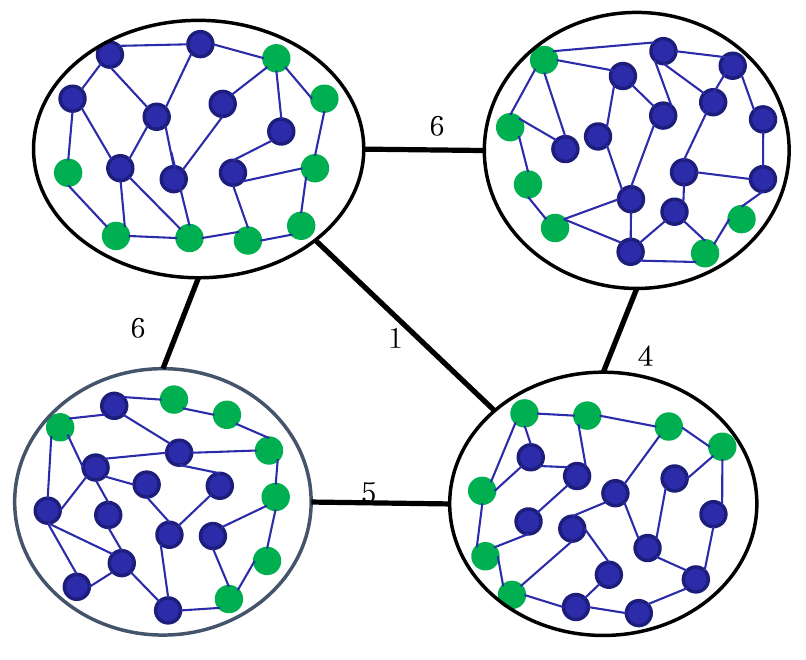}
\end{minipage}}
\subfigure[]{
\begin{minipage}{0.45\linewidth}
\includegraphics[width=1.0\linewidth]{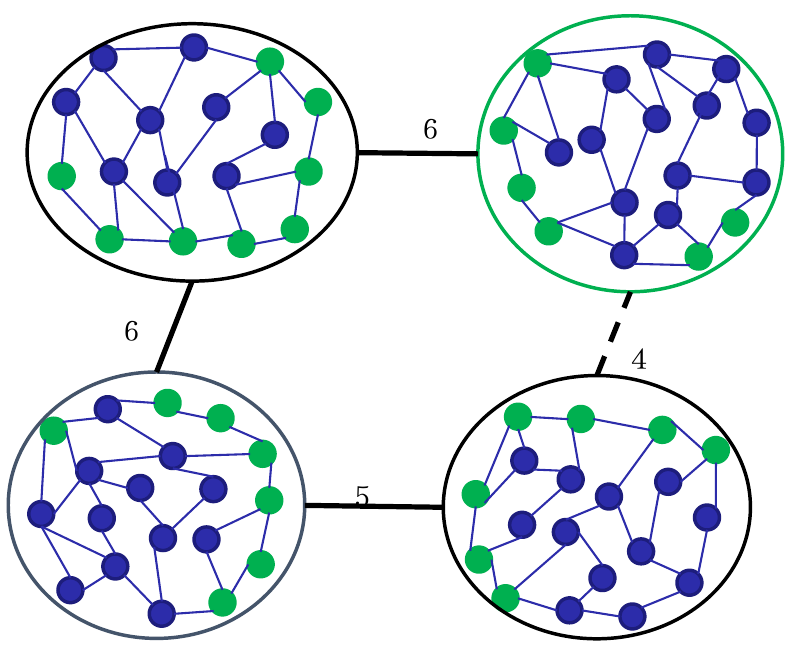}
\end{minipage}}
\subfigure[]{
\begin{minipage}{0.45\linewidth}
\includegraphics[width=1.0\linewidth]{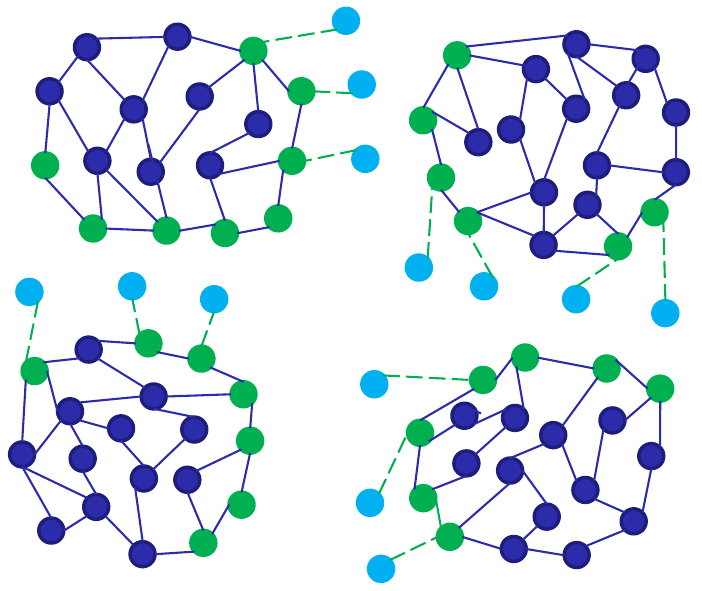}
\end{minipage}}
\caption{The image clustering procedure. (a). The images graph is grouped by NCut, where edges with weak connections are tend to be removed. (b). The cluster graph after graph cutting, where nodes are clusters and edges are lost edges in images graph, the number of lost edges are edge weights. (c). The solid lines represent the edges of constructed maximum spanning tree. The dotted line could be added to enhance connectivity of clusters. (d) The final expanded images clusters.}
\label{fig:image_clustering}
\end{figure}

The procedure of our image clustering algorithm is shown in Fig.\ref{fig:image_clustering}. In Fig. 2(a), The images graph is first grouped by using graph cut algorithm, where edges with weak connections tend to be removed. In Fig. 2(b), The cluster graph after graph cutting, where nodes are clusters and edges are lost edges in images graph, the number of lost edges are edge weights. In Fig. 2(c), the solid lines represent the edges of a constructed maximum spanning tree. The dotted line could be added to enhance the connectivity of clusters. Fig. 2(d) shows the final expanded image clusters. The complete image clustering algorithm is given in Alg.~\ref{alg:image_clustering}.

%%%%%%%%%%%%%%%%%%%%%%%%%%%%%%%%%%%%%%%%%% Algorithm %%%%%%%%%%%%%%%%%%%%%%%%%%%%%%%%%%%%%%%%%%
\begin{algorithm}
\caption{Image Clustering Algorithm}
\label{alg:image_clustering}
\begin{algorithmic}[1]
\Require an initial image graph $\mathcal{G} := \{(\mathcal{V}, \mathcal{E})\}$, maximum number of cluster size $\mathcal{S}_{\text{max}}$, completeness ratio $\mathcal{C}$, number of overlapped images between two clusters $\mathcal{O}_{\text{max}}$, number of images $n$.
\Ensure image clusters with intersection $\mathcal{G}_{\text{inter}} = \{\mathcal{G}_k\} $

\State $K := \lfloor n / \mathcal{S}_{\text{max}} \rfloor$ 
\State $\mathcal{G}_{\text{intra}} := $ GraphPartition($\mathcal{G}$), $\mathcal{G}_{\text{intra},i} \in \mathcal{G}_{\text{intra}}$, $i \in [K]$
\State Collect lost edges $E_{lost} := \{ \mathcal{E}_{k_1 k_2} | k_1, k_2 \in [K] \} $
\State Build cluster graph $\mathcal{G}_{\text{cluster}} := \emptyset$ from $E_{lost}$
\State $E_{\text{mst}} := \Kruskal(\mathcal{G}_{\text{cluster}})$, $i := 0, j := 0$, $\mathcal{G}_{\text{inter}} := \mathcal{G}_{\text{intra}}$
\While{$i < |E_{\text{mst}}|$}
    \State edge $\mathcal{E}_{k_1 k_2, i} = E_{\text{mst}, i}$ 
    \State $\mathcal{G}_{\text{inter}}^{k_1}, \mathcal{G}_{\text{inter}}^{k_2}$ add lost edges in $\mathcal{E}_{k_1 k_2, i}$
    \State $i \leftarrow i + 1$
\EndWhile

\While{completeness constraint not satisfied}
    \State Select edge $\mathcal{E}_{k_1 k_2, r}$ from $E_{lost}$ randomly
    \State $\mathcal{G}_{\text{inter}}^{k_1}, \mathcal{G}_{\text{inter}}^{k_2}$ add lost edges in $\mathcal{E}_{k_1 k_2, r}$
\EndWhile

\end{algorithmic}
\end{algorithm}
%%%%%%%%%%%%%%%%%%%%%%%%%%%%%%%%%%%%%%%%%% End Algorithm %%%%%%%%%%%%%%%%%%%%%%%%%%%%%%%%%%%%%%%%%

%------------------------------------------------------------------------
\subsection{Graph-based Local Reconstructions Merging}

After image clustering, each cluster can be reconstructed using a local SfM approach. Due to the robustness to outliers, we choose incremental SfM. As the reconstructed images are bounded below a threshold, the drift problem is alleviated. When all clusters are reconstructed, we need a final step to stitch them, as each cluster has its local coordinate frame. 

To construct a robust merging algorithm, we consider three main problems:
\begin{itemize}
    \item A cluster should be selected as the reference frame, which we referred to as \textbf{anchor node}.
    \item The merging step from other clusters to the anchor node should be as accurate as possible.
    \item As there may not exist overlap between anchor node and some other clusters, we have to find a path to merge them into an anchor node. Due to the accumulated errors, the path of each cluster to the anchor node shouldn't be too long.
\end{itemize}

To deal with the above problem, we construct a graph on the cluster level. The algorithm is composed of three main steps: (1) Cluster graph initialization. (2) Anchor node searching. (3) Path computation and simplification. For cluster graph initialization, we first find the common cameras between pairwise clusters, and compute the similarity transformations% by RANSAC \cite{DBLP:journals/cacm/FischlerB81} and Umeyama \cite{DBLP:journals/pami/Umeyama91}
. Then we build a MinST to select the most accurate edges. We found the anchor node by dealing with a minimum height tree (MHT) \cite{DBLP:journals/dam/LaberN04} problem. We first show how the problem can be constructed into an MinST problem.

\subsubsection{Pairwise Similarity Transformation}

We have discussed how to construct overlapping areas in Sec.~\ref{sec:image_clustering}, we further utilize the overlapped information to compute the pairwise similarity transformation. Given correspondences of camera poses, i.e., $\{P_{i_{k_1}}^{k_1}\}$ and $\{P_{i_{k_2}}^{k_2}\}$, we first estimate the relative scale. With relative scale known, the similarity estimation thus degenerated to euclidean estimation.

\paragraph{Relative Scale Estimation} To estimate relative scale, we need at least two points correspondences, $(C_{i_{k_1}}^{k_1}, C_{j_{k_1}}^{k_1})$ and $(C_{i_{k_2}}^{k_2}, C_{j_{k_2}}^{k_2})$, we can estimate the relative scale by
\begin{center}
\begin{equation}
    \hat{s}_{k_1 k_2}= (C_{i_{k_1}}^{k_1}-C_{j_{k_1}}^{k_1})/(C_{i_{k_2}}^{k_2}-C_{j_{k_2}}^{k_2}).
\end{equation}
\end{center}

As there may exists outliers, we choose $s_{k_1k_2}$ as
\begin{equation}
    s_{k_1k_2}=median\{\hat{s}_{k_1 k_2}\}.
\end{equation}

\paragraph{Euclidean Transformation Estimation} When the relative scale is known, the similarity transformation degenerates to euclidean estimation. That is, we only need to estimate the relative rotation and relative translation. Suppose a 3D point $X$ is located in global coordinate frame, and $X_1, X_2$ is located in local coordinate frame $k_1, k_2$ by two euclidean transformation $(R_1, t_1), (R_2, t_2)$ respectively. Then we have
\begin{equation}
    %\begin{split}
        X_1 = R_1 X + t_1,\quad X_2 = R_2 X + t_2.
    %\end{split}
\end{equation}

We can further obtain
\begin{equation}
    \begin{split}
        X_2 &= R_2 R_1^{-1}(X_ 1- t_1) + t_2\\
        &= R_2R_1^TX_1 + (t_2 - R_2 R_1^T t_1).
    \end{split}
\end{equation}

Then, the relative transformation is 
\begin{equation}
    R_{12} = R_2 R_1^T,\quad t_{12} = t_2-R_2R_1^Tt_1.
\end{equation}

Because cluster $k_1$ and $k_2$ are up to a scale $s_{k_1k_2}$, we should reformulate the relative transformation as
\begin{equation}
    \begin{split}
            t_{k_1k_2} &= t_{k_2} - R_{k_2}R_{k_1}^T t_{k_1}\\
               &= -R_{k_2}^T C_{k_2} + s_{k_1k_2} R_{k_2} R_{k_1}^T R_{k_1} C_{k_1}\\
               &= s_{k_1k_2}C_{k_1} - R_{k_2}^T C_{k_2},
    \end{split}
\end{equation}
where $C_{k_1}$ and $C_{k_2}$ are camera centers in cluster $k_1, k_2$ respectively. To handle the existence of outliers, we combined the euclidean estimation with RANSAC.

\subsubsection{Cluster Graph Initialization}

 In our approach, each cluster is deemed as a graph node, and the edges connect the nodes sharing some common cameras. Assume that there are $k$ clusters of cluster graph $\mathcal{C}$, and we denote the probability of obtaining a good transformation from the cluster pair $(i,j)$ as $p_{ij}$. Consequently, the probability that all $k-1$ edges in a spanning tree $c$ can be reconstructed is approximated as
\begin{equation}
\label{equ:probability}
    P = \prod_{i=1...k-1} p_{c_{k_1}c_{k_2}},
\end{equation}
where $c_{k_1}$ and $c_{k_2}$ are two clusters associated with the $k$-th edge of $c$. $P$ can be considered as the probability that a global 3D reconstruction can be reached provided all spanning pairs are correctly reconstructed. Then we try to maximize the probability $P$ defined in Equ.(\ref{equ:probability}). This is equivalent to minimizing the cost function
\begin{equation}
    f = -\mathop{\log}\left( \prod_{i=1...k-1} p_{c_{k_1}c_{k_2}} \right).
\end{equation}

To solve for the optimal spanning tree, we define the weight of an edge connecting clusters $i$ and $j$ as
\begin{equation}
    w_{ij} = -\mathop{\log} \left( p_{ij} \right).
\end{equation}

Now the problem of maximizing the joint probability $P$ is converted to the problem of finding a MinST in an undirected graph. Note that in MinST computing, the concrete value of the edge weights does not matter but the order of them does. That is, A reasonably comparable strengths of connections between clusters, instead of an accurate estimate of $P$, is sufficient to help generate a good spanning tree. This observation leads us to the following residual-error weight definition scheme.

\paragraph{Residual Error}
As a reliable measure of the goodness of cluster merging, we use the Mean Square Distance (MSD) to help define the edge weight. The Mean Square Error (MSE) from cluster $k_1$ to cluster $k_2$ is defined as:

\begin{equation}
\label{equ:msd}
    \mse_{k_1 k_2} = \frac{1}{2n} \left( \sum_{i=1}^n \left\| T_{k_1,k_2} X_i^{k_1} - X_i^{k_2} \right\|_2^2 \right)^{\frac{1}{2}},
\end{equation}

where $T_{k_1, k_2}$ is the similarity transformation from cluster $k_1$ to cluster $k_2$, $X_i^{k}$ is the $i$-th common points in cluster $k$. Equ.(\ref{equ:msd}) describes the transformation error from cluster $k_1$ to cluster $k_2$. To convert MSE to a symmetric metric, we use the maximum of mse to define the MSD:
\begin{equation}
    \msd (k_1, k_2) = \mathop{\max} (\mse_{k_1k_2}, \mse_{k_1k_2}).
\label{equ:sym msd}
\end{equation}

Then the edge weight between vertices $k_1$ and $k_2$ in $C$ is defined $\msd(k_1,k_2)$.

\subsubsection{Miminum Height Tree Construction}

After computing all the weights, the graph initialization process has been completed. Then, we can construct an MinST by Kruskal algorithm to select the most accurate $N - 1$ similarity transformations. After finding an MinST, We need to find a base node as the reference of the global alignment of all clusters in the MinST. We impose restrictions on the selection of the base node: (1) The base node should be suitably large. (2) The path from the other nodes to the base node shouldn't be too long. The first constraint is considered for efficiency. The second constraint is used to avoid error accumulation. Taking a similar idea of \textbf{Minimum Height Tree (MHT)} in \cite{DBLP:journals/dam/LaberN04}, We convert the problem of finding the base node into an MHT problem. We first introduce the concept of MHT.

\begin{mht_def}
    For an undirected graph with tree characteristics, we can choose any node as the root. The resulting graph is then a rooted tree. Among all possible rooted trees, those with minimum height are called minimum height trees (MHTs).
\end{mht_def}

 We solve the MHT problem by merging the leaf nodes layer by layer. At each layer, we collect all the leaf nodes and merged them into their neighbors. At last, there may be two or one nodes left. If there are two nodes left, then we choose the node that has a larger size as the base node. If there is only one node left, then the node is the base node.  
 
\begin{figure*}
\centering
\includegraphics[width=0.9\linewidth]{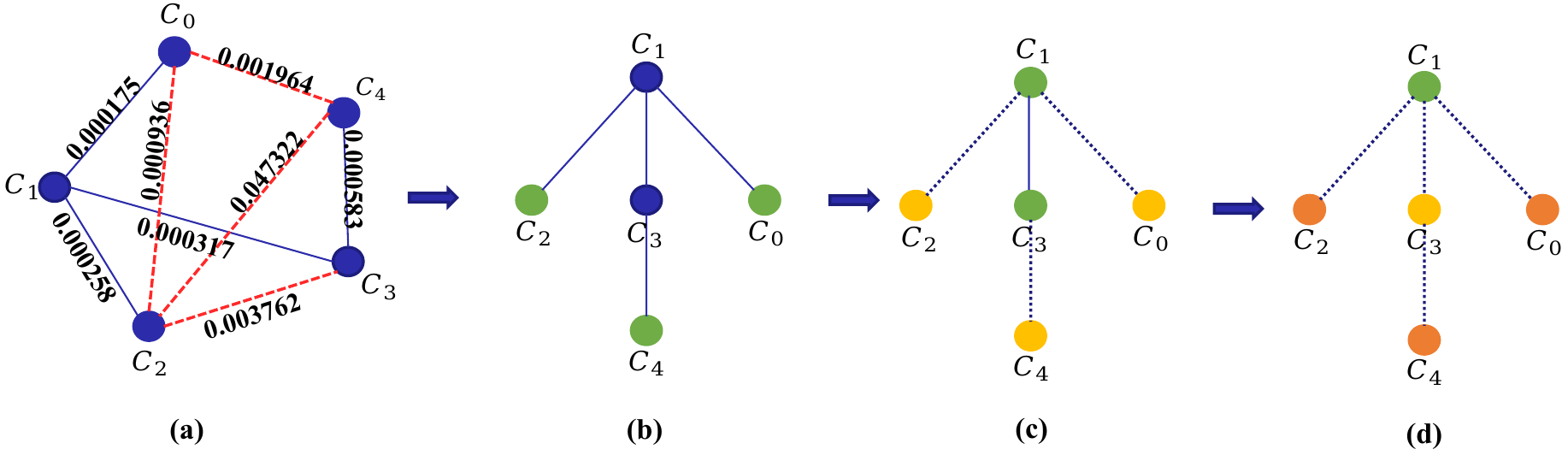}

\caption{The sub-reconstructions merging process. (a) shows the constructed MinST, where dotted lines represent the edges that have large $\msd$. In (b), leaf nodes($c_0, c_2, c_4$) are denoted by green, $c_0$ and $c_2$ are merged into $c_1$, $c_4$ is merged into $c_3$. In (c), the nodes that have been merged are marked by yellow, $c_1$ and $c_3$ are leaf nodes now. In (d), the leaf nodes which have been merged in the first layer are marked by dark yellow, and leaf nodes merged in second layer is marked by yellow. As there are two leaf nodes left, we choose the node with larger size (suppose $\left|c_1\right| > \left|c_3\right|$, then $c_3$ should be merged into $c_1$, and $c_1$ is the base node.)}
\label{fig:sub-model merging}
\end{figure*}

\begin{figure}
\centering

\includegraphics[width=1.0\linewidth]{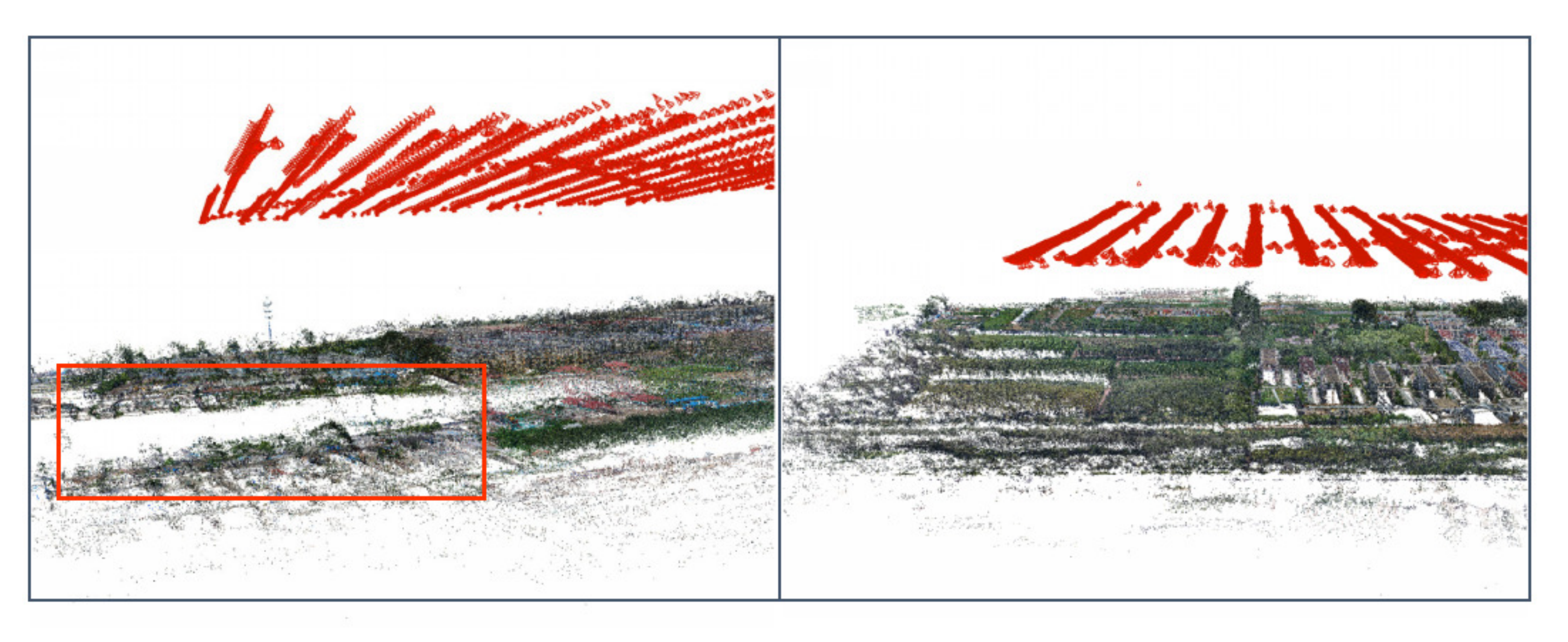}

\caption{The alignment results with and without our graph-based local reconstructions merging algorithm.}
\label{fig:with without}
\end{figure}

 The merging process is depicted by Fig.\ref{fig:sub-model merging}. The advantage of using algorithm to find the base node is depicted in Fig.\ref{fig:with without}. Owing to the robustness of our algorithm, which can find accurate similarity transformations and the edges which have large $\msd$ are filtered, we are able to merge all sub-reconstructions accurately. The full sub-model merging algorithm is illustrated in Alg.\ref{alg: sub-model merging algorithm}.

%%%%%%%%%%%%%%%%%%%%%%%%%%%%%%%%%%%%%%%% Figure %%%%%%%%%%%%%%%%%%%%%%%%%%%%%%%%%%%%%%%%%%%
%\begin{figure}
%\centering
%\includegraphics[width=0.5\linewidth]{img/pre_mst.png}

%\caption{The minimum spanning tree constructed from a cluster graph, the red dotted lines are deleted from the cluster graph.}
%\label{fig: mst construction}
%\end{figure}
%%%%%%%%%%%%%%%%%%%%%%%%%%%%%%%%%%%%%%%% End Figure %%%%%%%%%%%%%%%%%%%%%%%%%%%%%%%%%%%%%%%%%%%

%%%%%%%%%%%%%%%%%%%%%%%%%%%%%%%%%%%%%%%% Figure %%%%%%%%%%%%%%%%%%%%%%%%%%%%%%%%%%%%%%%%%%%

%%%%%%%%%%%%%%%%%%%%%%%%%%%%%%%%%%%%%%%%%%%%%%%%%algorithm%%%%%%%%%%%%%%%%%%%%%%%%%%%%%%%%%%%%%%%%%%%%%
\begin{algorithm}
\caption{Graph-Based Local Reconstructions Merging Algorithm}
\label{alg: sub-model merging algorithm}
\begin{algorithmic}[1]

\Require 
Clusters $\mathcal{C} := \{C_k\} $, Corresponding 3D points $\mathcal{X} := \{X_{ik}\ |\ i \in [n] \},
\mathcal{X^{'}} := \{X_{ik}^{'}\ |\ i \in [n]\}$, where $k \in [m]$

\Ensure 
The final merged cluster $\mathcal{C}_{\text{final}}$

\State Initialize Cluster Graph $\mathcal{G} := \{\mathcal{N}, \mathcal{E}\}$

\State Construct an MinST $\mathcal{M}$ by Kruskal algorithm from Cluster Graph $\mathcal{G}$

\State i := 0

\While{$|\mathcal{M}| > 1$}
    \State Find all leaf nodes $\{\mathcal{N}^i_{\text{leaf}}\}$ and their connected nodes $\{\mathcal{N}^i_{\text{con}}\}$ %and edges $\{\mathcal{E}^i_{lc}\}$ at i-th layer
    
    \State Replace $\{\mathcal{N}^i_{\text{con}}\}$ with $\{\mathcal{N}^i_{\text{leaf}} + \mathcal{N}^i_{\text{con}}\}$

    \State Remove %edges $\{\mathcal{E}^i_{lc}\}$ and 
    nodes $\mathcal{N}^i_{\text{leaf}}$
    
    \If{$|\mathcal{M}| = 2$} 
        \State Select the cluster that has a larger size as the base node
        \State Remove the other node
        \State break
    \EndIf
    
    \State i := i + 1
    
\EndWhile

\end{algorithmic}
\end{algorithm}

%------------------------------------------------------------------------
\section{Experiments}

In this section, we evaluate our GraphSfM on different kinds of datasets, including ambiguous datasets and large scale aerial datasets.

\subsection{Experimental Environments}
Our \emph{GraphSfM} algorithm is implemented based on COLMAP \cite{DBLP:conf/cvpr/SchonbergerF16}, and we use the incremntal pipeline of COLMAP to reconstruct local clusters. All the experiments are performed on a PC with 4 cores Intel 7700 CPU and 32GB RAM. Besides, we use SIFT \cite{DBLP:journals/ijcv/Lowe04} that was implemented by vlfeat\footnote{https://www.vlfeat.org/api/sift.html} to extract feature points for all the evaluated SfM approaches. 

\subsection{Datasets Overview}

To evaluate the robustness and efficiency of our algorithm, we first construct and collect some different kinds of datasets. The first kind of datasets are collected from 9 outdoor scenes, which include small scale and medium scale datasets, and the number of images is from 60 to 2248. The second kind of datasets are collected from public datasets, which include ourdoor scenes (Gerrard Hall, Person Hall, South Building) \cite{DBLP:conf/cvpr/SchonbergerF16} and ambiguous scenes (Stadium and Heaven Temple)~\cite{DBLP:conf/eccv/ShenZFZQ16}. The last kind of datasets are 3 large scale aerial datasets, where the memory requirement and efficiency are challenges for traditional approaches.

\subsection{Efficiency and Robustness Evaluation}
\begin{figure}
\centering
\includegraphics[width=1.0\linewidth]{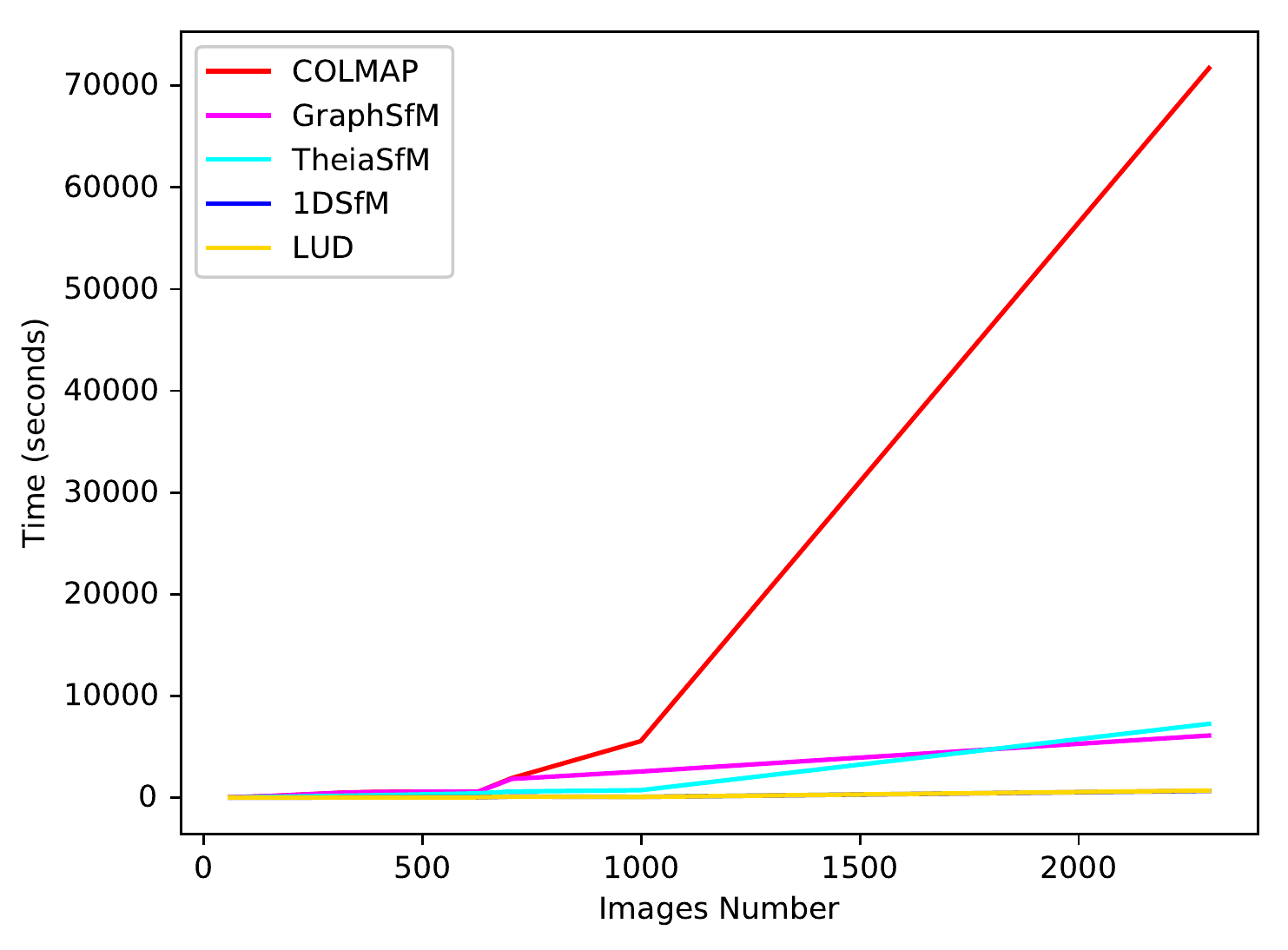}

\caption{Efficiency evaluation on datasets with different scales.}
\label{fig:time_eval}
\end{figure}

We evaluated the efficiency of our algorithm over 2 state-of-the-art incremental SfM approaches (TheiaSfM \cite{DBLP:conf/mm/SweeneyHT15} and COLMAP \cite{DBLP:conf/cvpr/SchonbergerF16}), and 2 state-of-the-art global SfM approaches (1DSfM~\cite{DBLP:conf/eccv/WilsonS14} and LUD~\cite{DBLP:conf/cvpr/OzyesilS15}). For sake of fairness, our GraphSfM runs on one computer, though it can run in a distributed mode. The evaluation results are shown in Fig.~\ref{fig:time_eval} and table~\ref{table:eff_eval}. It's not surprising that the incremental approaches take more time for reconstruction than global approaches. As the dataset scale increases, the time that is taken by COLMAP \cite{DBLP:conf/cvpr/SchonbergerF16} grows rapidly, due to the repetitive and time-consuming bundle adjustment \cite{DBLP:conf/iccvw/TriggsMHF99} step. Though our approach is a kind of incremental one, the scale of the images can be controlled to a constant size in each cluster. Thus the time of bundle adjustment can be highly reduced, and the time grows linearly as the number of images grows. Though TheiaSfM~\cite{DBLP:conf/mm/SweeneyHT15} is also an incremental SfM, it selects some good tracks~\cite{DBLP:journals/pr/CuiSH17} to perform bundle adjustment, which saves a lot of time but might become unstable in some cases. Besides, the time taken by TheiaSfM surpasses our GraphSfM when the scale of the image is over 2000. Table~\ref{table:eff_eval} gives more details of reconstruction results. It is obvious that our GraphSfM is as robust as COLMAP in terms of reconstructed cameras and is more accurate than other approaches in terms of reprojection errors. These facts illustrate the superior performance of our GraphSfM to handle large scale datasets. We emphasize that our algorithm just run on one computer and the reconstruction time could be reduced largely if we run it on more computers in distributed manner.

%%%%%%%%%%%%%%%%%%%%%%%%%%%%%%%%%%%%% table %%%%%%%%%%%%%%%%%%%%%%%%%%%%%%%%%%%%
\begin{table*}[]
\centering
\caption{Efficiency and accuracy evaluation with datasets which have different scales. $N_p, N_c, T$ represent the number of 3D points, the number of recovered cameras and the reconstruction time, respectively. Err represents the reprojection error and the best results are highlighted by bold font.}
\label{table:eff_eval}
\resizebox{\textwidth}{!}{
    \begin{tabular}{| c | c || c | c | c | c || c | c | c | c || c | c | c | c || c | c | c | c || c | c | c | c |} %表格列 全部居中显示
        \hline
 
        \multirow{2}{*}{\textbf{dataset}} & \multirow{2}{*}{\textbf{Images}} & 
        \multicolumn{4}{c||}{\textbf{COLMAP} \cite{DBLP:conf/cvpr/SchonbergerF16}} & 
        \multicolumn{4}{c||}{\textbf{TheiaSfM} \cite{DBLP:conf/mm/SweeneyHT15}} &
        \multicolumn{4}{c||}{\textbf{1DSfM} \cite{DBLP:conf/eccv/WilsonS14}} &
        \multicolumn{4}{c||}{\textbf{LUD} \cite{DBLP:conf/cvpr/OzyesilS15}} & 
        \multicolumn{4}{c|}{\textbf{Ours}} \\
        
        \cline{3-22} & \ & $N_c$ & $N_p$ & Err & $T$
                         & $N_c$ & $N_p$ & Err & $T$ 
                         & $N_c$ & $N_p$ & Err & $T$ 
                         & $N_c$ & $N_p$ & Err & $T$ 
                         & $N_c$ & $N_p$ & Err & $T$ \\
        \hline

        DS-1  & 60 
        & 60 & 16387 & 0.48 & 26.22
        & 60 & 8956 & 1.92 & 10.93
        & 60 & 8979 & 1.92 & 1.32
        & 60 & 8979 & 1.92 & 1.36 
        & 60 & 13923 & \textbf{0.46} & 24.48 \\
        \hline
        
        DS-2 & 158
        & 158 & 68989 & 0.42 & 170.34
        & 158 & 39506 & 1.91 & 87.71
        & 157 & 39527 & 1.92 & 7.76
        & 158 & 39517 & 1.92 & 7.95
        & 158 & 62020 & \textbf{0.44} & 168.48 \\
        \hline
        
        DS-3  & 214
        & 214 & 71518 & 0.512 & 122.64
        & 138 & 6459 & 1.70 & 45.89
        & 187 & 7080 & 1.54 & 4.25
        & 162 & 5099 & 1.45 & 1.69
        & 214 & 68882 & \textbf{0.49} & 121.56 \\
        \hline
        
        DS-4  & 319
        & 319 & 154702 & 0.50 & 529.14
        & 204 & 11550 & 1.80 & 186.08
        & 290 & 142967 & 1.82 & 17.53
        & 270 & 13484 & 1.76 & 18.77
        & 319 & 151437 & \textbf{0.47} & 482.40 \\
        \hline
        
        DS-5  & 401
        & 370 & 166503 & 0.58 & 568.68
        & 305 & 23742 & 1.97 & 241.61
        & 348 & 23081 & 1.88 & 18.89
        & 316 & 22160 & 1.85 & 17.93
        & 370 & 164495 & \textbf{0.55} & 562.74 \\
        \hline
 
        DS-6  & 628 
        & 628 & 268616 & 0.39 & 562.74
        & 628 & 133300 & 1.92 & 421.23
        & 610 & 133146 & 1.91 & 34.80
        & 628 & 133747 & 1.91 & 35.03
        & 628 & 259333 & \textbf{0.39} & 605.58 \\
        \hline
        
        DS-7  & 704 
        & 703 & 345677 & 0.58 & 1918.86
        & 449 & 35659 & 1.86 & 603.85
        & 641 & 42716 & 1.93 & 108.15
        & 547 & 34296 & 1.91 & 97.82
        & 703 & 346394 & \textbf{0.55} & 1839.90 \\
        \hline
        
        DS-8  & 999 
        & 980 & 419471 & 0.52 & 1918.86
        & 733 & 40246 & 1.86 & 731.84 
        & 745 & 172864 & 1.77 & 77.80
        & 611 & 31254 & 1.74 & 70.78
        & 980 & 416512 & \textbf{0.50} & 2570.34 \\
        \hline
        
        DS-9  & 2248 
        & 2248 & 1609026 & \textbf{0.63} & 71736 
        & 2248 & 187392 & 2.47 & 7255.70
        & 2247 & 188102 & 2.48 & 667.59
        & 2248 & 188134 & 2.47 & 694.74
        & 2242 & 1445227 & 0.65 & 6108.06 \\
        \hline
    \end{tabular}
}

\end{table*}

\subsection{Evaluation on Public Datasets}
\begin{figure*}
\centering

\includegraphics[width=1.0\linewidth]{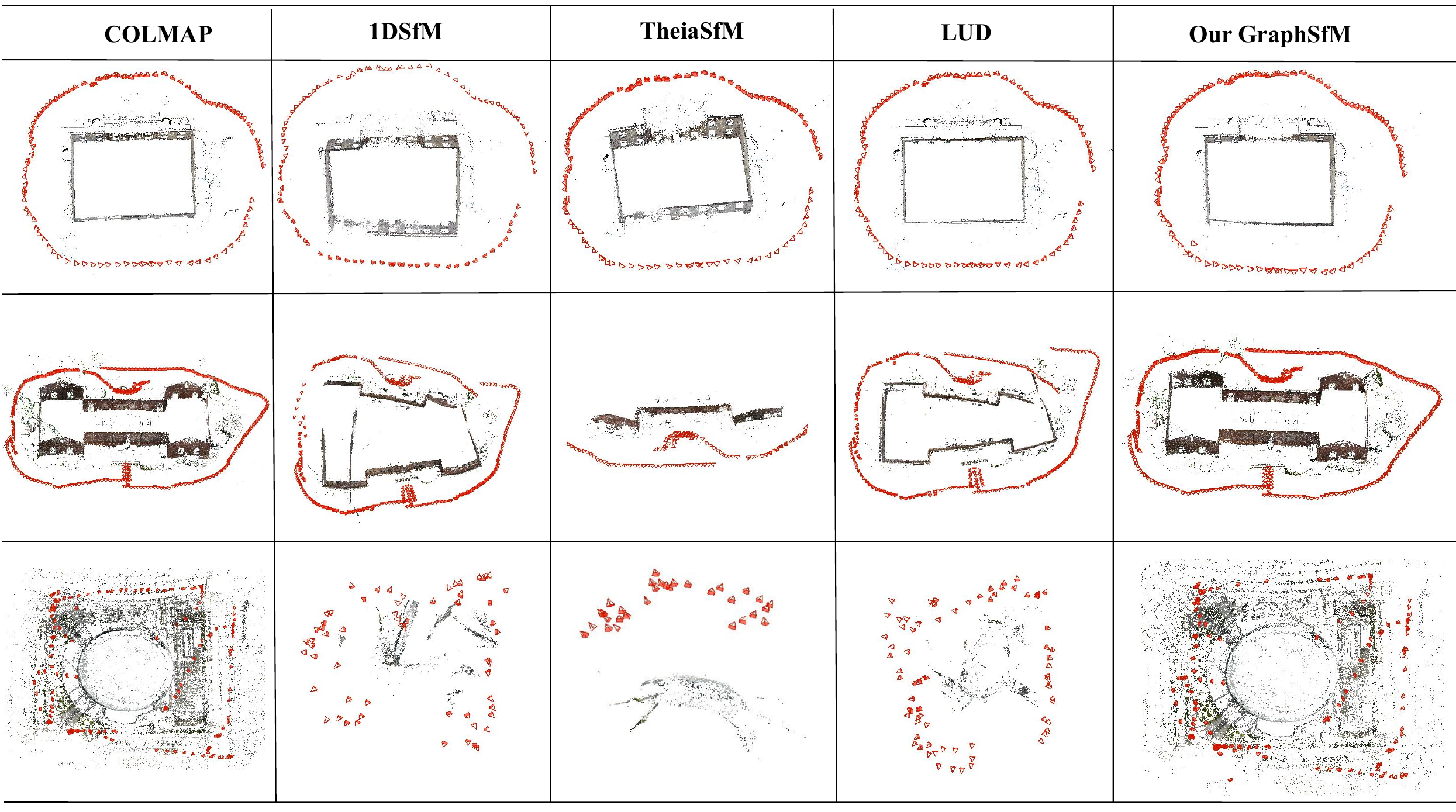}

\caption{Reconstruction results on temple public datasets. From top to bottom are respectively Gerrard Hall, Person Hall and Guangzhou Stadium datasets.}
\label{fig:public_datasets}
\end{figure*}

We evaluated our algorithm on several public datasets \cite{DBLP:conf/cvpr/SchonbergerF16, DBLP:conf/eccv/ShenZFZQ16}. For these small scale datasets, we only run our GraphSfM on one computer. Some visual results are shown in Fig.~\ref{fig:public_datasets} and statistics are given in table~\ref{table:statistics}. COLMAP again is the most inefficient approach, and our approach is 1.2 - 3 times faster than COLMAP, though we only run it by one computer. TheiaSfM selects good tracks~\cite{DBLP:journals/pr/CuiSH17} for optimization and the two global approaches are most efficient. However, as shown in Fig.~\ref{fig:public_datasets}, we can see that both the global approaches failed in Person Hall and Guangzhou Stadium datasets, which shows global approaches are easily disturbed by outliers. As an incremental approach, TheiaSfM also failed in Person Hall and Guangzhou Stadium datasets. Our approach is as robust as COLMAP, however, more efficient than it.

%%%%%%%%%%%%%%%%ng ma%%%%%%%%%%%%%%%%%% Figure %%%%%%%%%%%%%%%%%%%%%%%%%%%%%%%%%%%%%%%%%%%

\begin{figure}
\centering
\subfigure[COLMAP \cite{DBLP:conf/cvpr/SchonbergerF16}]{
    \begin{minipage}{0.35\linewidth}
        \includegraphics[width=1.0\linewidth]{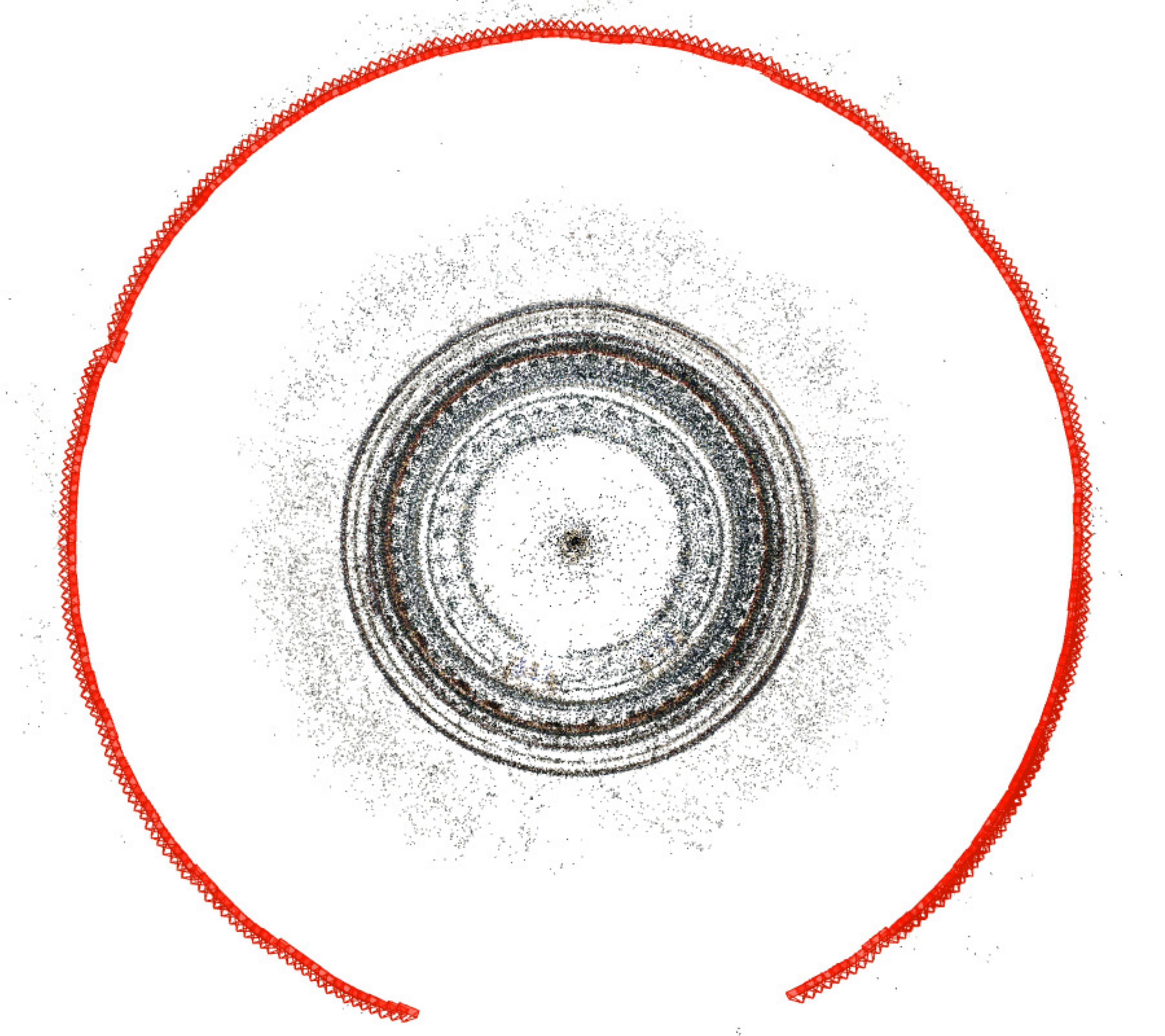}
    \end{minipage}
    \begin{minipage}{0.5\linewidth}
        \includegraphics[width=1.0\linewidth]{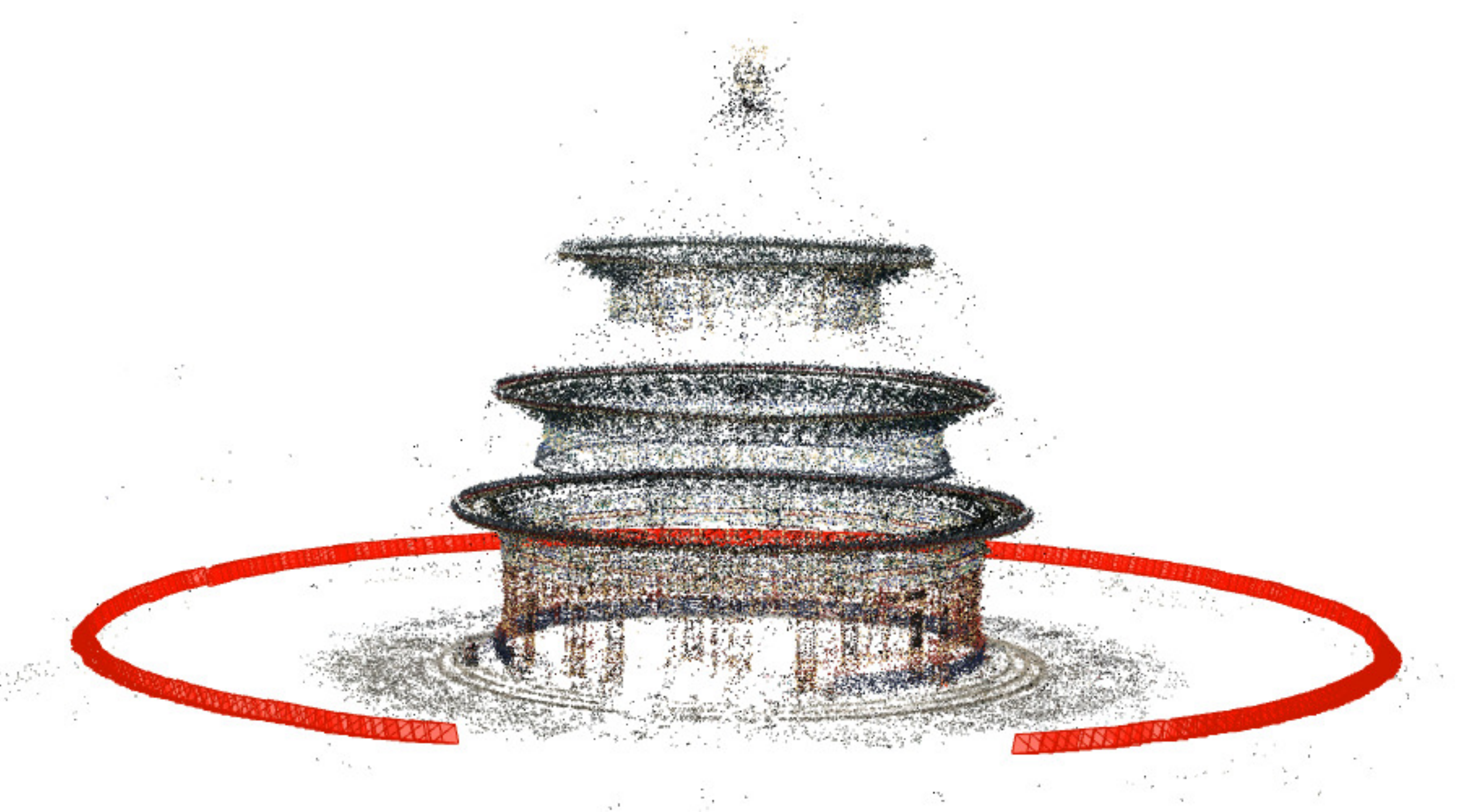}
    \end{minipage}
}
\subfigure[LUD~\cite{DBLP:conf/cvpr/OzyesilS15}]{
    \begin{minipage}{0.35\linewidth}
        \includegraphics[width=1.0\linewidth]{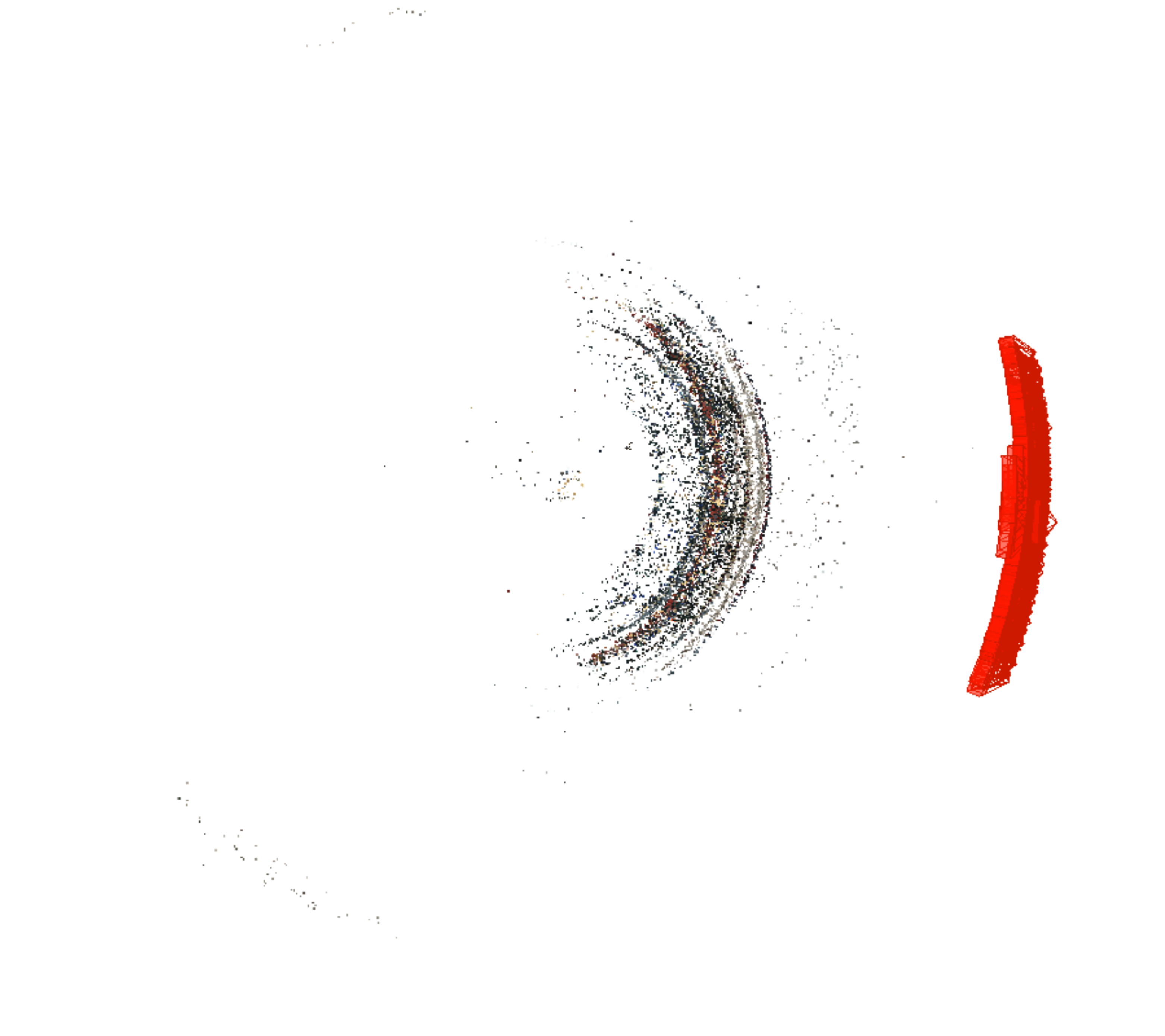}
    \end{minipage}
    \begin{minipage}{0.5\linewidth}
        \includegraphics[width=1.0\linewidth]{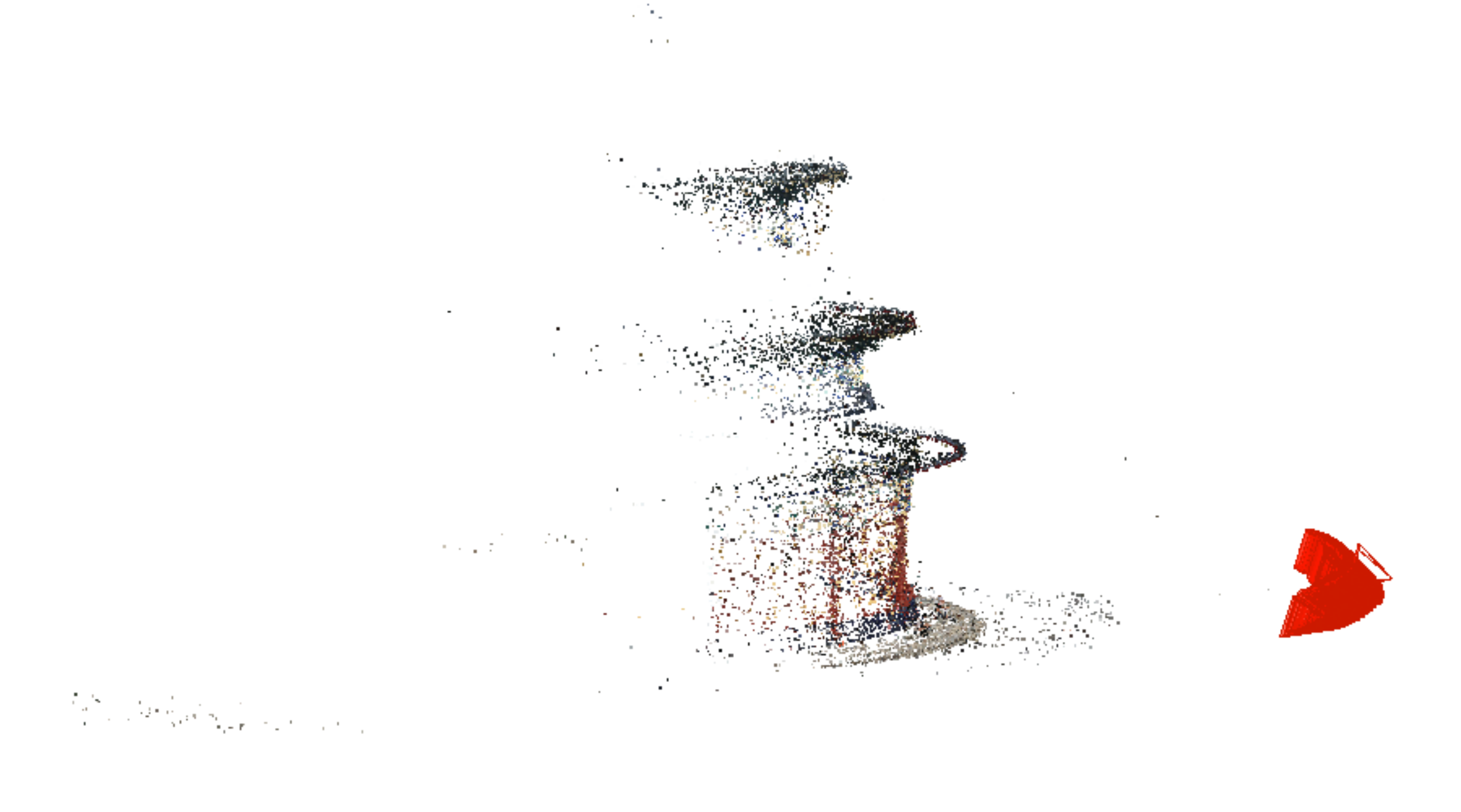}
    \end{minipage}
}
\subfigure[GraphSfM]{
    \begin{minipage}{0.35\linewidth}
        \includegraphics[width=1.0\linewidth]{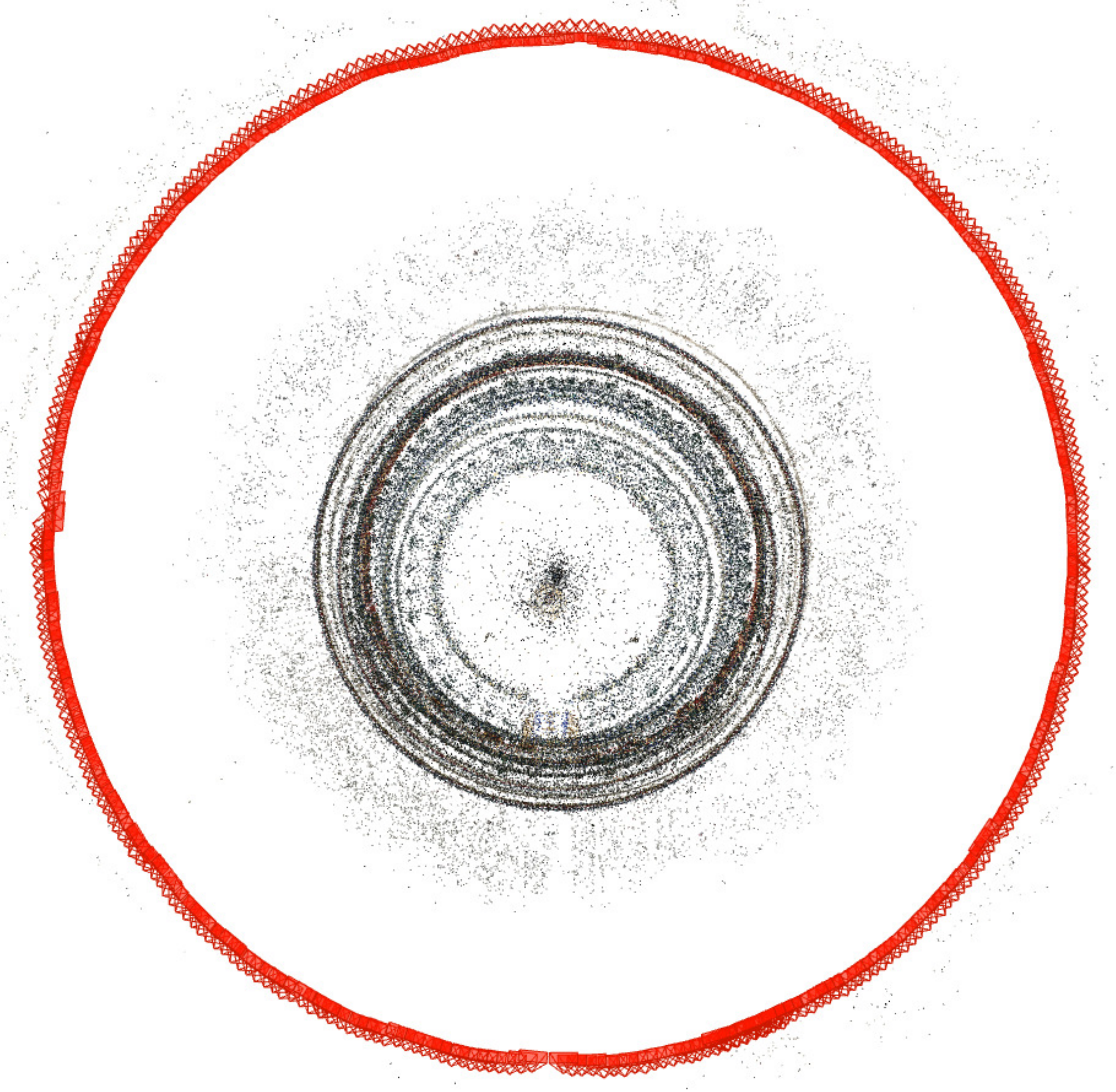}
    \end{minipage}
    \begin{minipage}{0.48\linewidth}
        \includegraphics[width=1.0\linewidth]{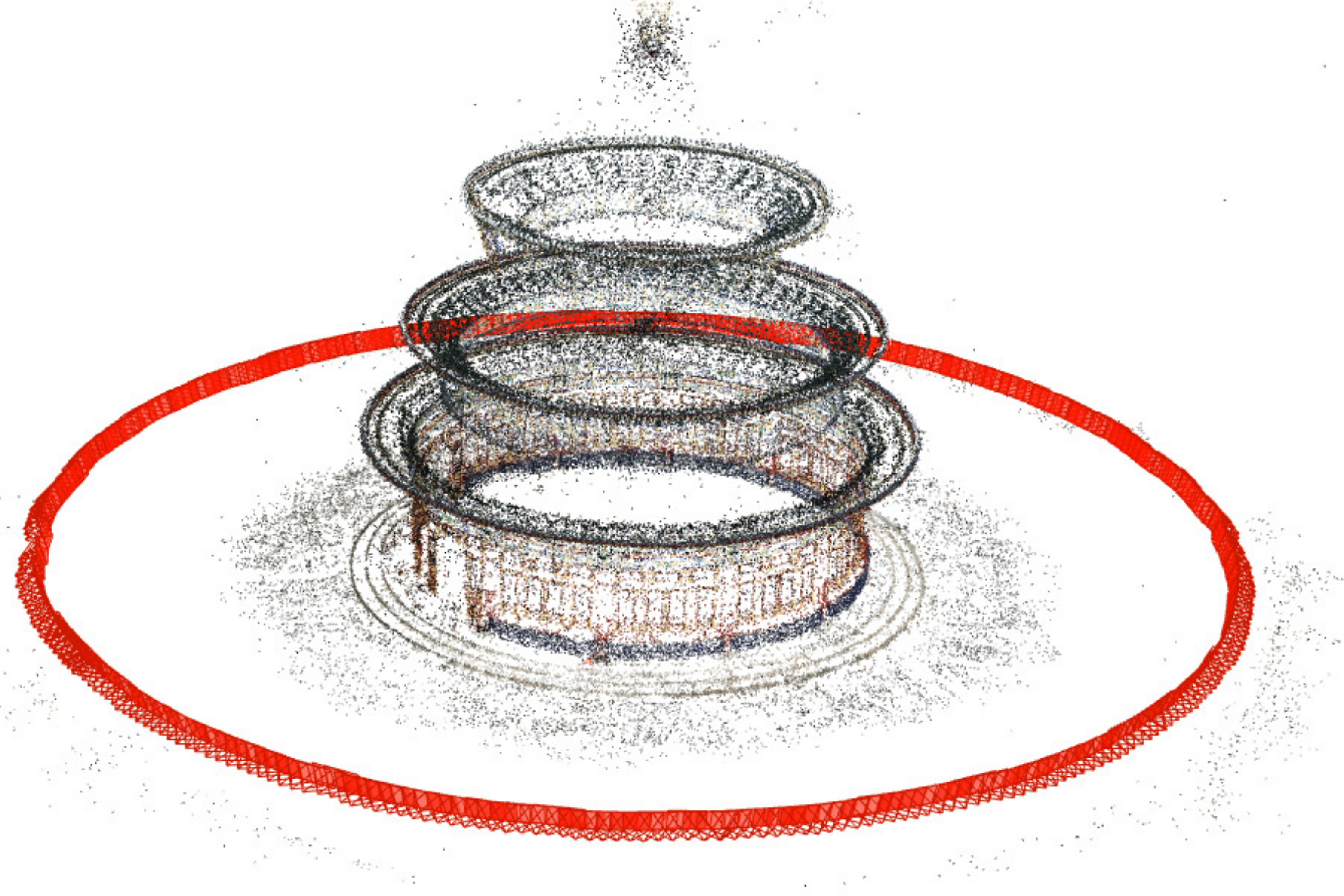}
    \end{minipage}
}
\caption{Reconstructions on temple heaven dataset. (a): The result reconstructed by COLMAP \cite{DBLP:conf/cvpr/SchonbergerF16}. (b): The result reconstructed by LUD~\cite{DBLP:conf/cvpr/OzyesilS15}. (c): The result reconstructed by our GraphSfM.}
\label{fig:temple_heaven}
\end{figure}

\paragraph{Ambiguous Datasets} It's a challenging work to reconstruct on ambiguous datasets for SfM approaches. Though feature matches are filtered by geometric constraints, there are still lots of wrong matches pass the verification step. As is shown in Fig.\ref{fig:temple_heaven}, our GraphSfM shows advantages to traditional SfM approaches in this kind of datasets. Due to the image clustering step, some wrong edges are discarded in clusters, thus the individual reconstructions are not affected by the wrong matches. However, it is hard to detect wrong matches in traditional SfM approaches, especially in self-similar datasets or datasets with repeated structures, which is the major reason for the failure in ambiguous datasets.

%%%%%%%%%%%%%%%%%%%%%%%%%%%%%%%%%%%%% table %%%%%%%%%%%%%%%%%%%%%%%%%%%%%%%%%%%%
\begin{table*}[]
    \centering
        \caption{Comparison of reconstruction results. $N_p, N_c$ represents the number of 3D points and the number of recovered cameras, respectively. $T_{GC},T_{SfM},T_{PCA},T_{\sum}$ respectively denotes the time cost (seconds) of graph cluster step, local SfM step, point clouds alignment step, and the total time.}
    \label{table:statistics}
\resizebox{\textwidth}{!}{
    \begin{tabular}{| c | c || c | c | c || c | c | c || c | c | c || c | c | c || c | c | c | c | c | c |} %表格列 全部居中显示
        \hline
 
        \multirow{2}{*}{\textbf{dataset}} & \multirow{2}{*}{\textbf{Images}} & \multicolumn{3}{c||}{\textbf{COLMAP} \cite{DBLP:conf/cvpr/SchonbergerF16}} & \multicolumn{3}{c||}{\textbf{TheiaSfM} \cite{DBLP:conf/mm/SweeneyHT15}} &
        \multicolumn{3}{c||}{\textbf{1DSfM} \cite{DBLP:conf/eccv/WilsonS14}} &
        %\multicolumn{3}{c||}{\textbf{Jiang} \cite{}} &
        \multicolumn{3}{c||}{\textbf{LUD} \cite{DBLP:conf/cvpr/OzyesilS15}} & 
        \multicolumn{6}{c|}{\textbf{Ours}} \\
        
        \cline{3-20} & \ & $N_c$ & $N_p$ & $T_{\sum}$ & $N_c$ & $N_p$ & $T_{\sum}$ & $N_c$ & $N_p$ & $T_{\sum}$ & $N_c$ & $N_p$ & $T_{\sum}$ & 
        $N_c$ & $N_p$ & $T_{GC}$ & $T_{SfM}$ & %$T_{PCA}$ & 
        $T_{BA}$ & $T_{\sum}$ \\
        \hline

        Gerrard Hall  & 100 & \textbf{100}  & 42795  & 303.07
                            & \textbf{100}  & 50232 & 93.35 
                            & 99 & 49083 & 15.82
                            % &93  & 28965 & 17.926 & 
                            & \textbf{100} & 44844 & \textbf{13.82}
                            & \textbf{100} & 42274 & 0.01 & 114 & %0.02922 & 
                              3.85 & 118.68\\
        \hline
        
        Person Hall & 330 & \textbf{330} & 141629 & 1725.80
                          & 113 & 39101 & 157.42 
                          & 42 & 6239 & 768.75
                            %- & - & - & 
                          & 325 & 93386 & \textbf{107.56} 
                          & \textbf{330} & 140859 & 0.04 & 713.94 & %0.989017 & 
                                25.13 & 742.92 \\
        \hline
        
        South Building & 128 & \textbf{128} & 61151 & 303.06
                             & \textbf{128} & 68812 & 155.84 
                             & \textbf{128} & 436640 & \textbf{27.71}
                             % &  &  &  
                             & \textbf{128} & 69110 & 34.70 
                             & \textbf{128} & 58483 & 0.03 & 125.28 & %0.028326 & 
                                4.75 & 131.28 \\
        \hline
        
        Stadium & 157 & \textbf{157} & 85723 & 418.74
                      & 30 & 6345 & 18.73 
                      & 65 & 6319 & 5.56
                        % &  &  &
                      & 77 & 4549 & \textbf{4.74} 
                      & \textbf{157} & 71605 & 0.03 & 403.86 & %0.278672 & 
                        16.167 & 421.62 \\
        \hline
        
        Heaven Temple & 341 & \textbf{341} & 185750 & 8678.76 
                            & 336 & 1201 & 46039 
                            & 339 & 13356 & \textbf{40.96}
                            %- & - & - & 
                            & 340 & 14019 & 44.09 
                            & \textbf{341} & 181583 & 0.04 & 2784.38 & %0.808667 & 
                                46.737 & 2856.26  \\
        \hline
 
    \end{tabular}
}

\end{table*} 

%%%%%%%%%%%%%%%%%%%%%%%%%%%%%%%%%% Figure %%%%%%%%%%%%%%%%%%%%%%%%%%%%%%%%%%%%%%%%%%%

\subsection{Evaluation on Large Scale Aerial Datasets}

Our approach is also evaluated on large scale aerial datasets. We evaluated our algorithm both on one computer sequentially(each cluster is reconstructed one by one) and on three computers in parallel mode. The reconstruction results are given in table~\ref{table:aerial_statistics}. Our algorithm can recover the same number of cameras as well as COLMAP. Besides, when running on a single computer, our approach is about 6 times faster than COLMAP; when running on three computers in distributed manner, our algorithm is about 17 times faster than COLMAP. Our algorithm can be further accelerated if we own more computer resources. TheiaSfM is slightly slower than our approach, however, the the number reconstructed 3D points is one magnitude less than our approach. Though 1DSfM and LUD are still the most efficient, their robustness would meet challenge in large scale aerial datasets. 

%%%%%%%%%%%%%%%%%%%%%%%%%%%%%%%%%%%%% table %%%%%%%%%%%%%%%%%%%%%%%%%%%%%%%%%%%%
\begin{table*}[]
    \centering
        \caption{Comparison of reconstruction results. $N_p, N_c$ represents the number of 3D points and the number of recovered cameras, respectively. $T$ denotes the total time and $T_d$ denotes the total time that is evaluated in a distributed system.}
    \label{table:aerial_statistics}
\resizebox{\textwidth}{!}{
    \begin{tabular}{| c | c || c | c | c || c | c | c || c | c | c || c | c | c || c | c | c | c |} %表格列 全部居中显示
        \hline
 
        \multirow{2}{*}{\textbf{dataset}} & \multirow{2}{*}{\textbf{Images}} & \multicolumn{3}{c||}{\textbf{COLMAP} \cite{DBLP:conf/cvpr/SchonbergerF16}} & \multicolumn{3}{c||}{\textbf{TheiaSfM} \cite{DBLP:conf/mm/SweeneyHT15}} &
        \multicolumn{3}{c||}{\textbf{1DSfM} \cite{DBLP:conf/eccv/WilsonS14}} &
        \multicolumn{3}{c||}{\textbf{LUD} \cite{DBLP:conf/cvpr/OzyesilS15}} & 
        \multicolumn{4}{c|}{\textbf{Ours}} \\
        
        \cline{3-18} & \ & $N_c$ & $N_p$ & $T$ & $N_c$ & $N_p$ & $T$ & $N_c$ & $N_p$ & $T$ & $N_c$ & $N_p$ & $T$ & 
        $N_c$ & $N_p$ & $T$ & $T_d$ \\
        \hline
        
        Aerial-5155 & 5155 
        & \textbf{5155} & 1798434 & 41823.27
        & 4383 & 203942 & 3527.43
        & 4591 & 243490 & 482.45
        & 4723 & 278924 & \textbf{390.59}
        & \textbf{5155} & 1834875 %& 0.033 & 1348.56 & 903.28 
        & 2491.78 & 936.74 \\
        \hline
        
        Aerial-7500 & 7500 
        & \textbf{7455} & 5184368 & 95007.59
        & 5327 & 432347 & 6237.91
        & 6264 & 478234 & 931.84
        & 5934 & 467230  & \textbf{832.40}
        & \textbf{7455} & 4968142 %& 0.044 & 2784 & 2902.3 
        & 5834.37 & 2166.59  \\
        \hline
        
        Aerial-12306 & 12306 
        & \textbf{11259} & 3934391 & 146172
        & 8347 & 478237 &  25783.12
        & 8923 & 509543 &  4941.27
        & 8534 & 489238 & \textbf{4589.73}
        & \textbf{11259} & 3916724 %& 3.704 & 3947.1 & 14287 
        & 22663.86 & 8970.01  \\
        \hline
 
    \end{tabular}
}

\end{table*}

%%%%%%%%%%%%%%%%%%%% Figure %%%%%%%%%%%%%%%%%%%%%%%%%%%%%%%%%%%%%%%%%%%
\begin{figure*}
\centering

\includegraphics[width=1.0\linewidth]{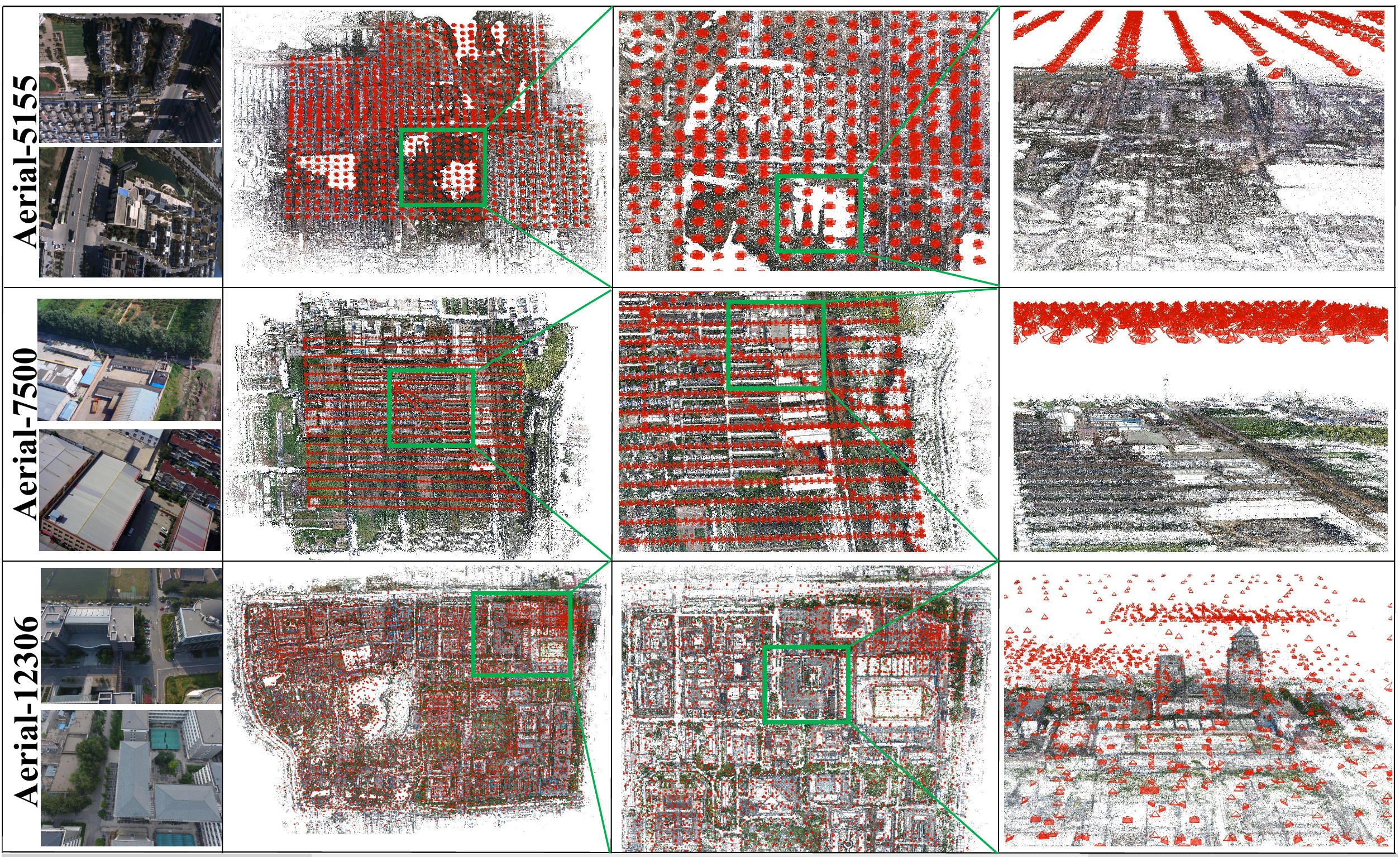}

\caption{Reconstructions on large scale aerial datasets.}
\label{figen dataset:aerial}
\end{figure*}

%------------------------------------------------------------------------
\section{Conclusion}
In this article, we proposed a new SfM pipeline called GraphSfM, which is based on graph theory, and we designed a unified framework to solve large scale SfM tasks. Our two steps graph clustering algorithm enhances the connections of clusters, with the help of a MaxST. In the final fusing step, the construction of MinST and MHT allows us to pick the most accurate similarity transformations and to alleviate the error accumulation. Thus, our GraphSfM is highly efficient and robust to large scale datasets, and also show superiority in ambiguous datasets when compared with traditional state-of-the-art SfM approaches. Moreover, GraphSfM can be implemented on a distributed system easily, thus the reconstruction is not limited by the scale of datasets.

%\section{Front matter}

%The author names and affiliations could be formatted in two ways:
%\begin{enumerate}[(1)]
%\item Group the authors per affiliation.
%\item Use footnotes to indicate the affiliations.
%\end{enumerate}
%See the front matter of this document for example. 
%You are recommended to conform your choice to the journal you 
%are submitting to.

%\appendix
%\section{My Appendix}
%Appendix sections are coded under \verb+\appendix+.

%\verb+\printcredits+ command is used after appendix sections to list 
%author credit taxonomy contribution roles tagged using \verb+\credit+ 
%in frontmatter.

\printcredits

\newpage

%% Loading bibliography style file
%\bibliographystyle{model1-num-names}
\bibliographystyle{cas-model2-names}

% Loading bibliography database
\bibliography{cas-refs}
%\bibliography{mybibfile}

%\vskip3pt

\end{document}